\documentclass{article}
\pdfoutput=1

\usepackage[preprint]{neurips_2026}

\usepackage[utf8]{inputenc}
\usepackage[T1]{fontenc}
\usepackage{hyperref}
\usepackage{url}
\usepackage{booktabs}
\usepackage{amsfonts}
\usepackage{amssymb}
\usepackage{mathtools}
\usepackage{graphicx}
\usepackage{subcaption}
\usepackage{nicefrac}
\usepackage{microtype}
\usepackage{xcolor}
\usepackage{tikz}
\usetikzlibrary{arrows.meta,positioning,calc,fit}

\definecolor{auditnavy}{HTML}{17365D}
\definecolor{auditblue}{HTML}{E3EEF8}
\definecolor{auditgreen}{HTML}{E7F3E8}
\definecolor{auditgold}{HTML}{F7ECD9}
\definecolor{auditgray}{HTML}{F1F2F4}

\newcommand{\queries}{\mathcal{Q}}
\newcommand{\support}{\nu}
\newcommand{\closure}{\mathcal{C}_{\queries,H,\support}}

\newcommand{\modelstate}{S_M}
\newcommand{\envop}{\mathcal{K}}
\newcommand{\modelop}{\widehat{\mathcal{K}}_M}

\title{The Intervention Gap in Latent World Models}

\author{%
  Donna Vakalis \\
  Mila -- Quebec AI Institute \\
  University of Montreal \\
  \texttt{donna.vakalis@olympian.org}
}

\begin{document}

\maketitle

\begin{abstract}
Planning-time intervention fidelity is a distinct, measurable property of a
learned world model: whether the model's own open-loop transitions move task
variables the way matched environment interventions do. In the
settings we test, it is neither revealed by reward fit nor ensured by
task-anchored training. Across released TD-MPC2 checkpoint sizes, episode
return falls as an operator-error diagnostic on task observables grows,
while reward-prediction error stays small and nearly flat, and a
self-supervised world model trained without task signal preserves the same
operator substantially better than a task-anchored model on the shared task.
A capture-gated matched-intervention audit then localizes what fails. On
Cheetah, three LeWorldModel checkpoints capture the current task query and
support decodable real intervention effects; however, their imagined
five-step effects are worse than predicting no effect and worse than an
environment-endpoint oracle. The failure is task-direction rotation with
excess gain, not feature collapse. This severe pattern is conditional: five PreJEPA seeds
retain an oracle-relative deficit without it, Finger Spin experiments extend
the deficit beyond locomotion with heterogeneous severity across seeds, and
shared-bank effect geometry is both candidate- and support-dependent. We
also test practice-side questions. In DreamerV3 the posterior distribution,
not its
sample, carries the current query; ensemble disagreement ranks error only
near training support; and a frozen support-aware score degrades held-out
error ranking in both tested transfer directions while native disagreement
remains informative in both. We conclude that intervention fidelity must be
audited directly, capture-first, on the model's native interface.
\end{abstract}

\section{Introduction}
\label{sec:introduction}

A world model used for planning must do two different things: represent the
current variables that matter to the task, and predict how candidate actions
change them \citep{sutton1991dyna,ha2018worldmodels}. Standard training
signals certify neither requirement separately. One-step prediction quality
correlates poorly with control performance \citep{lambert2020objective}, and
a model can pass next-step diagnostics while its implicit transition
structure is incoherent \citep{vafa2024evaluating}. A task variable can be
readable from a real encoded state even when the model assigns the wrong
consequence to an intervention; conversely, a transition score is difficult
to interpret when the relevant quantity was never recoverable from the model
state in the first place.

This paper develops one claim across five sets of results.
\emph{Planning-time intervention fidelity, the property that a model's own
open-loop transitions move task variables the way matched environment
interventions do, is distinct and measurable. Value-based
training signals neither reveal nor ensure it; it can fail severely while
reward fit, current-state capture, and overall feature alignment all look
fine; and it therefore has to be audited directly, capture-first, on the
model's native interface.} Every empirical statement below is relative to a
declared task query, physical horizon, intervention support, and native model
interface; none is a population-of-architectures claim.

The argument proceeds in three steps. First, value-based signals neither
measure intervention fidelity nor produce it. On a released TD-MPC2
checkpoint-size sweep,
episode return falls as an operator-error diagnostic on task observables
grows, while reward-prediction error stays small and nearly flat, and the
operator diagnostic ranks checkpoints differently from Bellman-residual and
value-slice metrics that agree with each other
(Section~\ref{sec:valuechannel}). Nor does training against task signal
secure the property: on a shared task, a self-supervised world model trained
with no task signal preserves the same operator substantially better than a
task-anchored model. The value channel neither detects the failure nor, on
this comparison, prevents it.

Second, a capture-gated matched-intervention audit localizes what fails
(Sections~\ref{sec:audit}, \ref{sec:dissociation}, and
\ref{sec:boundaries}). On Cheetah, three LeWorldModel checkpoints capture the
current task query and support decodable real intervention effects, while
their own imagined five-step effects are worse than predicting no effect at
all and worse than an environment-endpoint oracle. The failure is not
wholesale feature collapse: the complete predicted feature change remains
strongly aligned with the real one while its task component is rotated to
near-orthogonal and amplified. The severe pattern is conditional rather than
universal. Five independently trained PreJEPA predictors retain an
oracle-relative deficit without reproducing the severe conjunction; Finger
Spin extends the oracle-relative gap beyond locomotion with heterogeneous
severity across seeds; and on a shared evaluation, action-effect geometry is
candidate- and support-dependent rather than an architecture ranking.

Third, the same program yields practical guidance for auditing and trusting
such models (Section~\ref{sec:practice}). In a DreamerV3 study, the
categorical posterior distribution is the sole tested surface that supports
the current query across checkpoints; its sample, its mode, and the
deterministic state all fail. Ensemble disagreement ranks error near training
support and deteriorates away from it, and a frozen support-aware
augmentation of disagreement degrades error ranking on held-out families in
both tested transfer directions, while architecture-native disagreement
remains informative in both.

The contributions, in the order the argument uses them, are:
(i)~released-checkpoint evidence that reward fit can be silent about, and
task-anchored training insufficient for, operator fidelity on task
observables; (ii)~a capture-gated matched-intervention audit that separates
target availability, real-effect resolvability, and fidelity of the model's
own transitions; (iii)~a prospective Cheetah/LeWorldModel dissociation of
current-query capture from severe propagation distortion, characterized as
task-direction rotation with excess gain; (iv)~boundary results (a PreJEPA
non-replication of the severe conjunction, heterogeneous Finger severity, and
support-dependent shared-bank geometry) that scope the phenomenon without
erasing its oracle-relative deficit; and (v)~interface and uncertainty
results, including a bidirectional held-out falsifier for a frozen
support-aware score, that turn the audit into usable practice.

\section{Two Audits of Planning-Time Fidelity}
\label{sec:audit}

Let \(x\) denote a Markov state or a declared sufficient observation history.
Fix a family of task queries \(\queries=\{q_1,\ldots,q_m\}\), a physical
horizon \(H\), and a distribution \(\support\) over roots and action
sequences. For an action \(a\), define the environment operator
\begin{equation}
    (\envop_a q)(x)
    =
    \mathbb{E}\!\left[
        q(X_{t+1})
        \mid X_t=x,\operatorname{do}(A_t=a)
    \right],
\end{equation}
and let \(\modelop\) denote the corresponding operator induced by candidate
\(M\)'s own open-loop transition on its planner-exposed state \(\modelstate\).
Every diagnostic statement in this paper is relative to
\((\queries,H,\support)\) and to the model's native interface. Open-loop
latent rollouts are evaluated because they are the object planners and
imagined-rollout learners consume
\citep{hafner2019planet,hafner2020dreamer,hansen2022tdmpc}.

We use two instruments of increasing resolution.

\subsection{Operator error on task observables}
\label{sec:operatordiag}

The coarser instrument asks how far \(\modelop\) is from \(\envop\) on task
observables, without asking why. Ridge probes with grouped cross-validation
decode each query from the model's latent state; the model then advances its
latent through its own native prediction mode to the planning-time horizon,
and the probe-decoded outcome is compared with the same query evaluated on
the real environment outcome under the same actions. The result is a single
operator-error score per candidate on the declared query family. Probe
quality is reported separately from operator error, so a poorly decodable
observable is not silently charged to the transition. This diagnostic is
deliberately complementary to value-based metrics: it shares their
task-relativity but scores the transition on task observables rather than on
reward, value, or return \citep{farahmand2017vaml,grimm2020value}.

The coarse score conflates two failure sources: the information may be absent
from the latent state, or present but moved wrongly by the transition.
Section~\ref{sec:valuechannel} uses the coarse score, where that conflation
does not matter for the conclusions drawn; the rest of the paper uses the
finer instrument, which separates the sources.

\subsection{The capture-gated matched-intervention audit}
\label{sec:gatedaudit}

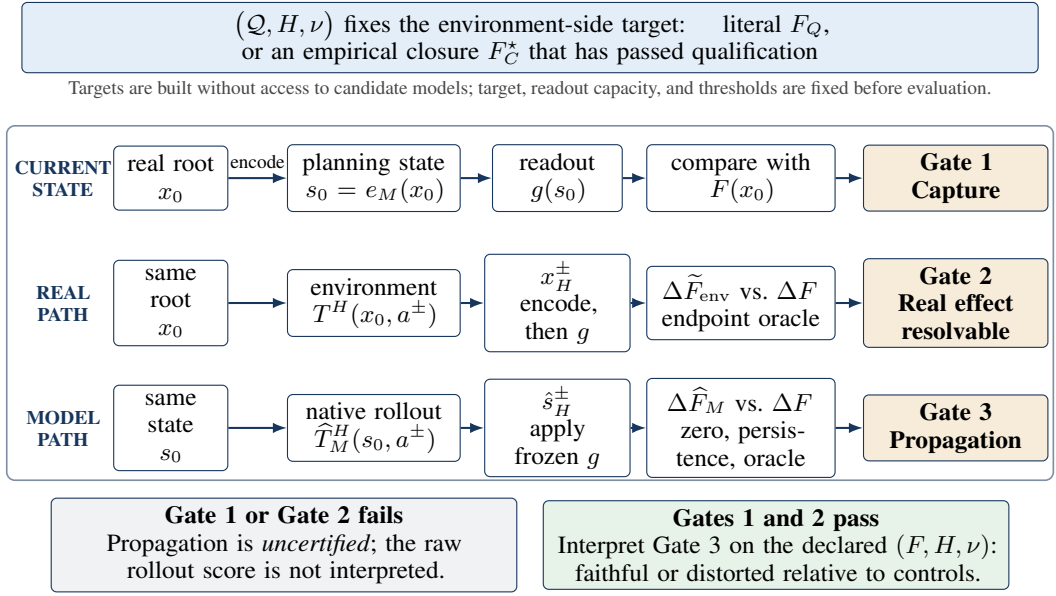
\begin{figure}[t]
\centering
\resizebox{\linewidth}{!}{\begin{tikzpicture}[
    x=1cm,
    y=1cm,
    font=\small,
    flow/.style={-{Latex[length=1.8mm]},line width=0.75pt,draw=auditnavy},
    box/.style={draw=auditnavy,rounded corners=2pt,fill=white,align=center,
                minimum height=0.82cm,inner sep=3pt},
    process/.style={box},
    target/.style={box},
    gate/.style={box,fill=auditgold,text width=2.15cm,font=\small\bfseries},
    rowlabel/.style={font=\scriptsize\bfseries,text=auditnavy,align=center,
                    text width=1.25cm},
    smallnote/.style={font=\scriptsize,align=center,text=black!72}
]
\node[box,fill=auditblue,text width=12.8cm,minimum height=0.9cm] (spec) at (6.6,3.05)
{\(\bigl(\mathcal{Q},H,\nu\bigr)\) fixes the environment-side target:
\quad literal \(F_Q\), or an empirical closure \(F_C^\star\) that has passed qualification};
\node[smallnote] at (6.6,2.38)
{Targets are built without access to candidate models; target, readout capacity, and thresholds are fixed before evaluation.};

\node[rowlabel] (currentlabel) at (0.62,1.25) {CURRENT\\STATE};
\node[box,text width=1.25cm] (rootcapture) at (2.0,1.25) {real root\\\(x_0\)};
\node[process,text width=2.0cm] (statecapture) at (4.6,1.25)
{planning state\\\(s_0=e_M(x_0)\)};
\node[target,text width=1.45cm] (readcapture) at (6.95,1.25)
{readout\\\(g(s_0)\)};
\node[box,text width=2.2cm] (comparecapture) at (9.3,1.25)
{compare with\\\(F(x_0)\)};
\node[gate] (gatecapture) at (12.05,1.25)
{Gate 1\\Capture};

\draw[flow] (rootcapture) -- node[above,font=\scriptsize,inner xsep=1pt]{encode} (statecapture);
\draw[flow] (statecapture) -- (readcapture);
\draw[flow] (readcapture) -- (comparecapture);
\draw[flow] (comparecapture) -- (gatecapture);

\node[rowlabel] (reallabel) at (0.62,-0.35) {REAL\\PATH};
\node[box,text width=1.25cm] (rootreal) at (2.0,-0.35) {same root\\\(x_0\)};
\node[process,text width=2.0cm] (envroll) at (4.6,-0.35)
{environment\\\(T^H(x_0,a^\pm)\)};
\node[target,text width=1.65cm] (realendpoint) at (6.95,-0.35)
{\(x_H^\pm\)\\encode, then \(g\)};
\node[box,text width=2.2cm] (comparereal) at (9.3,-0.35)
{\(\Delta\widetilde F_{\mathrm{env}}\) vs.\ \(\Delta F\)\\endpoint oracle};
\node[gate] (gatereal) at (12.05,-0.35)
{Gate 2\\Real effect resolvable};

\draw[flow] (rootreal) -- (envroll);
\draw[flow] (envroll) -- (realendpoint);
\draw[flow] (realendpoint) -- (comparereal);
\draw[flow] (comparereal) -- (gatereal);

\node[rowlabel] (modellabel) at (0.62,-1.95) {MODEL\\PATH};
\node[box,text width=1.25cm] (rootmodel) at (2.0,-1.95) {same state\\\(s_0\)};
\node[process,text width=2.0cm] (modelroll) at (4.6,-1.95)
{native rollout\\\(\widehat T_M^H(s_0,a^\pm)\)};
\node[target,text width=1.65cm] (modelendpoint) at (6.95,-1.95)
{\(\hat s_H^\pm\)\\apply frozen \(g\)};
\node[box,text width=2.2cm] (comparemodel) at (9.3,-1.95)
{\(\Delta\widehat F_M\) vs.\ \(\Delta F\)\\zero, persistence, oracle};
\node[gate] (gatemodel) at (12.05,-1.95)
{Gate 3\\Propagation};

\draw[flow] (rootmodel) -- (modelroll);
\draw[flow] (modelroll) -- (modelendpoint);
\draw[flow] (modelendpoint) -- (comparemodel);
\draw[flow] (comparemodel) -- (gatemodel);

\node[box,fill=auditgray,text width=5.75cm,minimum height=0.95cm] (uncertified) at (3.45,-3.45)
{\textbf{Gate 1 or Gate 2 fails}\\Propagation is \emph{uncertified}; the raw rollout score is not interpreted.};
\node[box,fill=auditgreen,text width=5.75cm,minimum height=0.95cm] (interpretable) at (9.75,-3.45)
{\textbf{Gates 1 and 2 pass}\\Interpret Gate 3 on the declared \((F,H,\nu)\): faithful or distorted relative to controls.};

\draw[draw=auditnavy!55,rounded corners=3pt,line width=0.6pt]
(-0.1,1.92) rectangle (13.3,-2.62);
\end{tikzpicture}}
\caption{\textbf{Capture-gated matched-intervention audit.}
A query family, physical horizon, and intervention support define an
environment-side target before candidate evaluation. Gate 1 tests whether the
target is readable from held-out real planning states. Gate 2 tests whether
the frozen endpoint readout resolves the real effect of the matched actions.
Only when both prerequisites pass is Gate 3 interpreted: the model's native
imagined effect is compared with the real effect, zero effect or persistence,
and the environment-endpoint oracle. Failure of either prerequisite yields
uncertified propagation rather than evidence of faithful propagation.}
\label{fig:capture_gated_audit}
\end{figure}

The finer instrument decomposes fidelity into three separately testable
prerequisites, summarized in Figure~\ref{fig:capture_gated_audit}.

\paragraph{Capture.}
A prespecified readout is trained on real encoded states and evaluated on
held-out roots. Capture reports coordinatewise and aggregate error under
development-only normalization.

\paragraph{Real-effect resolvability.}
Before model propagation is interpreted, the matched environment intervention
must produce an effect that the frozen endpoint readout can resolve at real
environment endpoints. If this check fails, the corresponding propagation
comparison is not interpreted.

\paragraph{Propagation.}
From the same real root, an action sequence is executed in the environment
and through the model's own open-loop transition. The frozen readout maps the
imagined endpoint to the target coordinates. Model error is compared with
persistence, with zero effect, and with the error of the real environment
endpoint under the same readout (the environment-endpoint oracle).

The gate order follows the structure of finite-horizon query error, which
obeys the schematic decomposition
\begin{equation}
\|\widehat q_h-q_h\|
\leq
\|B_h\|P_{0:h}\epsilon_{\mathrm{cap},0}
+
\sum_{t=0}^{h-1}\|B_h\|P_{t+1:h}
\left(\epsilon_{\mathrm{op},t}+\tau_{C,t}\right)
+
\tau_{q,h},
\end{equation}
where \(\epsilon_{\mathrm{cap},0}\) is capture error at the real root,
\(\epsilon_{\mathrm{op},t}\) is closure-restricted transition error,
\(\tau\)-terms are approximation residuals, and
\(P_{s:h}=\prod_{j=s}^{h-1}L_j\) collects environment stability factors
(Appendix~\ref{app:decomposition}). A propagation score is interpretable only
when the capture term is small; the decomposition identifies measured error
sources and is not a guarantee about realized return.

\paragraph{Matched interventions and supports.}
Each environment family supplies repeated roots and action contrasts. The
same action sequence, mapped into each model's action interface, is applied
to the environment and to every compatible candidate. Task-policy and
environment-random supports remain separate groups; roots, actions, horizons,
and model outputs are repeated measurements within a family. The action
contrasts test action sensitivity in the spirit of controllability metrics
for generative world models \citep{bruce2024genie,arai2024actbench}, but
against executed environment endpoints rather than alternative model
rollouts.

\paragraph{Isolation, readouts, and inference.}
Closure development, closure validation, probe training, calibration, and
evaluation use disjoint environment-family identities; candidate models are
unavailable during environment-side target construction, and evaluation data
affect no target rank, readout capacity, normalization, threshold, candidate
inclusion, or action support. The primary readout is a standardized ridge
probe with grouped nested cross-validation; controls include label and
feature shuffles, horizon-zero identity, persistence, executed-environment
endpoints, wrong-action and action-null contrasts, probe-capacity
sensitivity, and direct predicted-versus-real feature alignment. The complete
environment family is the uncertainty unit; checkpoints, roots, actions,
coordinates, and imagined samples are never treated as independent
replications, and each candidate is reported separately.

\paragraph{Calibration and closure boundary.}
The audit is calibrated on synthetic systems before application to learned
models: ten prespecified acceptance cases require it to distinguish capture
failure from propagation failure, reject persistence as successful dynamics,
recover a known stable rank, remain invariant to orthogonal basis changes,
and halt on split leakage; all ten pass under the fixed configuration
(Appendix~\ref{app:synthetic}). Learned-model results use the literal queries
\(F_Q\). A larger environment-side closure target
\(\closure=\operatorname{span}\{\envop_{a_0}\cdots\envop_{a_{t-1}}q\}\) is
used only after passing rank, null, and stability checks; the original
Cheetah construction did not pass those checks, and the qualified successor
holds only on matched action shells
(Appendix~\ref{app:closure_boundary}).

\section{The Value Channel Neither Reveals Nor Ensures Operator Fidelity}
\label{sec:valuechannel}

This section uses the coarse operator diagnostic of
Section~\ref{sec:operatordiag} to establish the limitation that motivates the
finer audit. Both results are single-environment observations on Cheetah
locomotion; they flag a failure mode rather than prove a general law.

\subsection{Return falls with operator error where reward fit is silent}
\label{sec:scissors}

\begin{figure}[t]
\centering
\includegraphics[width=\linewidth]{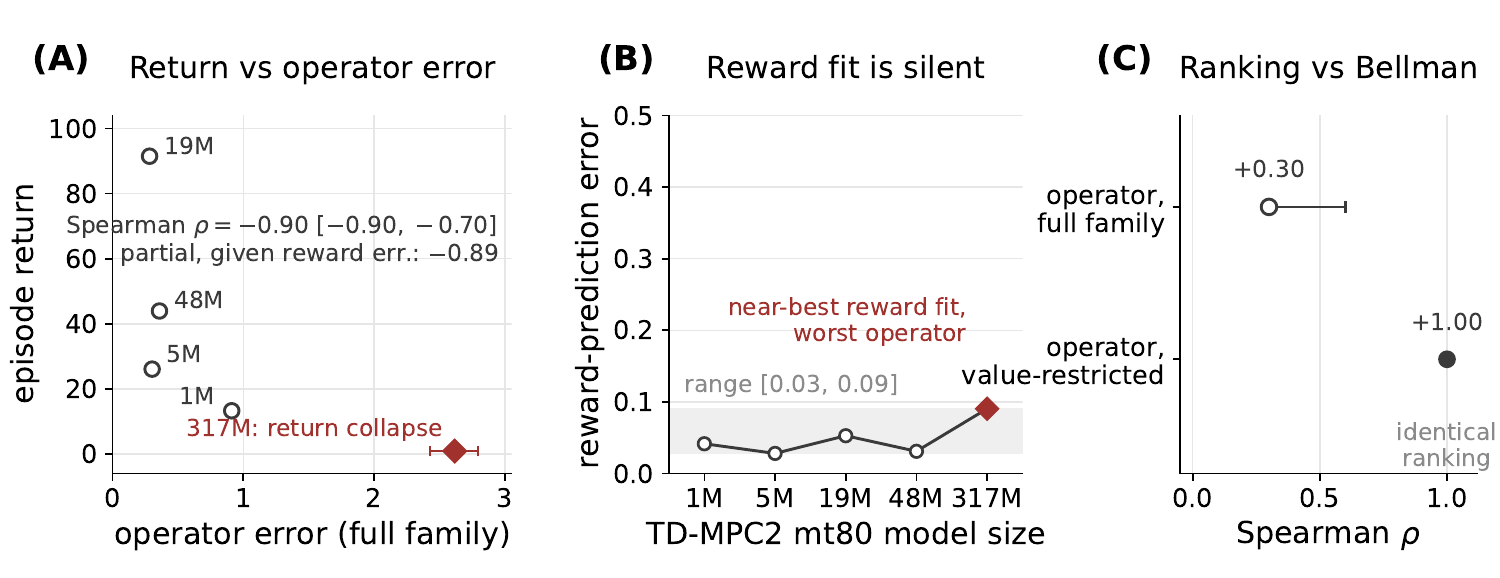}
\caption{\textbf{Reward fit is silent where the operator fails.}
Released multi-task TD-MPC2 checkpoints at five sizes on cheetah-run
(dependent sizes; descriptive sweep).
(A)~Episode return against operator error on task observables at the
planning-time horizon; rank correlation \(-0.90\) (anchor-bootstrap 95\%
interval \([-0.90,-0.70]\); leave-one-out range \(\geq-0.80\)); the
association survives controlling for reward-prediction accuracy (partial
rank correlation \(-0.89\)). The 317M checkpoint is marked: worst operator
error, collapsed return.
(B)~Reward-prediction error across the same checkpoints, flat within
\([0.03,0.09]\), with the same checkpoint marked: near-best reward fit,
worst operator.
(C)~Checkpoint rankings: full-observable operator error versus a
Bellman-residual metric (rank correlation \(+0.30\), 95\% interval
\([+0.30,+0.60]\)), and a value-restricted operator variant versus the same
Bellman residual (\(+1.00\)). One environment and \(n=5\) dependent sizes;
correlational evidence only.}
\label{fig:scissors}
\end{figure}

The released multi-task TD-MPC2 checkpoints \citep{hansen2024tdmpc2} provide
five model sizes trained under one recipe. Across this sweep, episode return
on cheetah-run falls as operator error on task observables at the
planning-time horizon grows, with rank correlation \(-0.90\), and the
association survives controlling for reward-prediction accuracy. Reward-prediction error
itself is small and nearly flat across the same checkpoints
(Figure~\ref{fig:scissors}). The largest checkpoint is the sharpest instance:
its reward fit is among the best in the sweep while its operator error is the
worst and its planning return collapses. On this sweep, the value channel's
own fit statistic is silent exactly where the transition moves task
observables wrongly.

The operator diagnostic also measures something the standard model-quality
proxies do not. Ranking the same checkpoints by operator error on the full
observable family agrees only weakly with a Bellman-residual ranking, while a
value-restricted variant of the operator score reproduces the
Bellman-residual ranking almost exactly
(Figure~\ref{fig:scissors}, right panel). This disagreement is itself
informative: restricted to value-relevant directions, the operator score
reproduces the value-channel ranking, so its additional signal comes from
the task-observable directions the value channel does not constrain. We
therefore treat the operator score as a complement to value-based metrics,
to be reported alongside them, not as a replacement
\citep{farahmand2017vaml,grimm2020value,lambert2020objective}.

These are five dependent released sizes from one training recipe on one
task; the sweep supports a correlational flag, not a selection rule or a
return guarantee.

\subsection{Task-anchored training did not confer operator fidelity}
\label{sec:anchoring}

\begin{figure}[t]
\centering
\includegraphics[width=\linewidth]{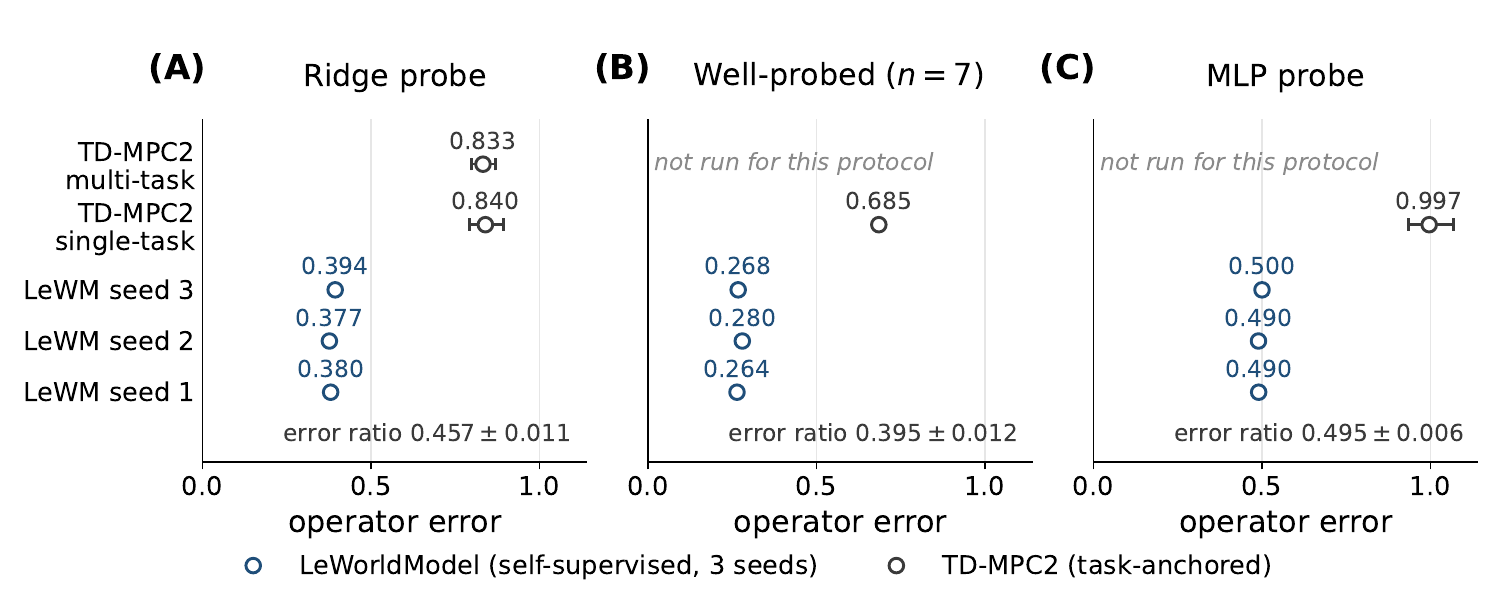}
\caption{\textbf{Task anchoring did not confer operator fidelity.}
Operator error on shared task observables at the planning-time horizon;
blue marks the self-supervised LeWorldModel seeds throughout.
(A)~Ridge probes on the full shared family: three independently seeded
LeWorldModel runs (\(0.384\pm0.009\)) versus task-anchored TD-MPC2
(single-task \(0.840\); multi-task \(0.833\)); error ratio
\(0.457\pm0.011\).
(B)~The family restricted to the \(n=7\) observables both models probe
well, against the single-task anchored checkpoint (the multi-task variant
was not run under this protocol): the gap grows (ratio \(0.395\pm0.012\)).
(C)~One-hidden-layer MLP-probe ablation, same comparator: seed-level 95\%
intervals remain disjoint (ratio \(0.495\pm0.006\)).
Two model families, one anchored seed per variant, differing input
modalities and native prediction modes; a bound on the anchoring claim, not
an architecture ranking.}
\label{fig:anchoring}
\end{figure}

Reward fit does not reveal operator error; a remaining possibility is that
task-anchored training nevertheless prevents it. On this comparison it did
not. Three independently seeded
LeWorldModel runs \citep{maes2026lewm}, self-supervised video-prediction
world models trained from scratch on cheetah-run with no task signal, are
compared with a task-anchored TD-MPC2 model trained on the same task, using
matched ridge probes on shared task observables and each model's native
prediction mode at the planning-time horizon. The self-supervised models
preserve the operator substantially better: their operator error is roughly
\(0.46\) of the task-anchored model's, with disjoint 95\% intervals across
all three seeds (Figure~\ref{fig:anchoring}).

Three control analyses support the comparison. A single-task and a
multi-task anchored checkpoint give the same answer, ruling out a multi-task
handicap. Restricting the observable family to coordinates both models probe
well makes the gap slightly larger, not smaller, ruling out probe asymmetry.
The seed-level intervals remain disjoint under a one-hidden-layer MLP probe
in place of the ridge probe. The comparison still spans only two model
families with one anchored seed, on models with different input modalities
and native prediction modes; it bounds the claim ``task anchoring secures
operator fidelity,'' and supports nothing stronger.

\subsection{Limits of the coarse diagnostic}
\label{sec:seam}

Together the two results say the value channel neither detects operator
error nor, on this comparison, prevents it. But the coarse score cannot say
what fails: an operator-error number conflates information that is absent
from the latent state with information that is present and moved wrongly,
and it aggregates over a support without matching interventions. The
remainder of the paper applies the capture-gated audit of
Section~\ref{sec:gatedaudit}, which separates those readings. The separation
matters concretely: the self-supervised family that preserves the coarse
operator better in this section is the same family whose matched intervention
effects fail most severely in the next. The two instruments measure
different properties of the same models; that is the reason to keep both,
and the reason neither raw score is comparable across instruments.

\section{Capture Without Propagation}
\label{sec:dissociation}

The capture-gated audit localizes the failure that the coarse diagnostic
cannot separate: a model can represent the task information and still
mispredict how actions change it. Both tasks in this and the next section come from the DeepMind Control
Suite \citep{tassa2018dmc}; Table~\ref{tab:capture_propagation_summary}
summarizes the three prospective rosters.

\begin{table}[t]
\centering
\caption{Capture-gated matched-intervention outcomes. Each row is one
task--architecture roster. Candidates are not pooled, and raw scores are not
comparable across tasks. Intervals use 4{,}096 whole-family bootstrap
resamples on \(n=96\) evaluation families per candidate. Severe means
predicted-effect \(R^2\) and its 95\% upper endpoint are both below zero.}
\label{tab:capture_propagation_summary}
\scriptsize
\begin{tabular}{llcccc}
\toprule
Task / model & Capture \(R^2\) & Real \(R^2\) & Predicted \(R^2\)
& Severe & Oracle deficit \\
\midrule
Cheetah / LeWM & 0.579--0.609 & 0.127--0.198 & \(-6.33\) to \(-3.36\)
& 3/3 & 3/3 \\
Cheetah / PreJEPA & pass & \(\approx 0.390\) & \(-0.005\) to \(0.064\)
& 0/5 & 5/5 \\
Finger / LeWM & 0.539--0.555 & 0.733--0.789 & \(-0.869\), \(0.287\), \(0.367\)
& 1/3 & 3/3 \\
\bottomrule
\end{tabular}
\end{table}

\subsection{The primary dissociation}
\label{sec:primary}

Three frozen LeWorldModel checkpoints \citep{maes2026lewm} are evaluated on
Cheetah task-policy support. Each evaluation family supplies three roots and
six matched actuator contrasts at a physical horizon of five environment
steps; the frozen target is endpoint torso speed, and readouts and thresholds
are fixed on 48 calibration families before each checkpoint is scored on 96
evaluation families.

All three checkpoints pass both prerequisites: the current query is readable
from real encoded roots, and the frozen readout resolves the real effects of
the matched pulses at real endpoints. Their own imagined effects then fail
both confirmatory comparisons. Predicted-effect \(R^2\) is far below
zero, worse than predicting no effect at all, with every whole-family 95\%
interval below zero, and every checkpoint's imagined endpoint is reliably
worse than the environment-endpoint oracle under the same readout
(Table~\ref{tab:capture_propagation_summary};
Figure~\ref{fig:intervention_gap_dissociation}, top row). Current-query
capture and real-effect resolvability therefore coexist with severe
distortion of the model's own five-step action effects, prospectively and
across the roster.

\subsection{Rotation with excess gain, not collapse}
\label{sec:geometryoffailure}

The simplest interpretation, that the transition loses the intervention
entirely, is not supported. In descriptive controls frozen with the confirmatory package, the
complete predicted feature change remains strongly aligned with the real one
(contrast cosine \(0.85\)--\(0.89\) across the three checkpoints), while the
same change projected onto the task readout is nearly orthogonal to the real
task effect and \(1.7\)--\(2.6\) times too large. Within the measured
surface, the failure is task-direction misalignment with excess gain: the
model moves its features broadly the right way and moves the task-relevant
component of that motion in the wrong direction, too strongly. This is why a
generic feature-space rollout error would have looked benign here, and why
trajectory-plausibility scores can miss the failure entirely
(Section~\ref{sec:related}). The characterization is descriptive; it
constrains candidate mechanisms without establishing one.

What the primary result does not establish is bounded in
Section~\ref{sec:limitations}: no absence of task information under every
nonlinear decoder, no claim at other horizons or supports, no planner-failure
or return claim, and no generalization beyond the tested roster, which is
the subject of the next section.

\section{The Deficit Replicates; Severe Distortion Does Not}
\label{sec:boundaries}

The severe Cheetah pattern could have been a universal property of latent
world models, a property of one architecture, or an artifact of one roster.
Three boundary results place it: the oracle-relative deficit reappears in
every tested roster, while the severe conjunction is conditional on model
family, task, and seed.

\subsection{A prospective non-replication in PreJEPA}
\label{sec:prejepa}

\begin{figure}[t]
\centering
\includegraphics[width=\linewidth]{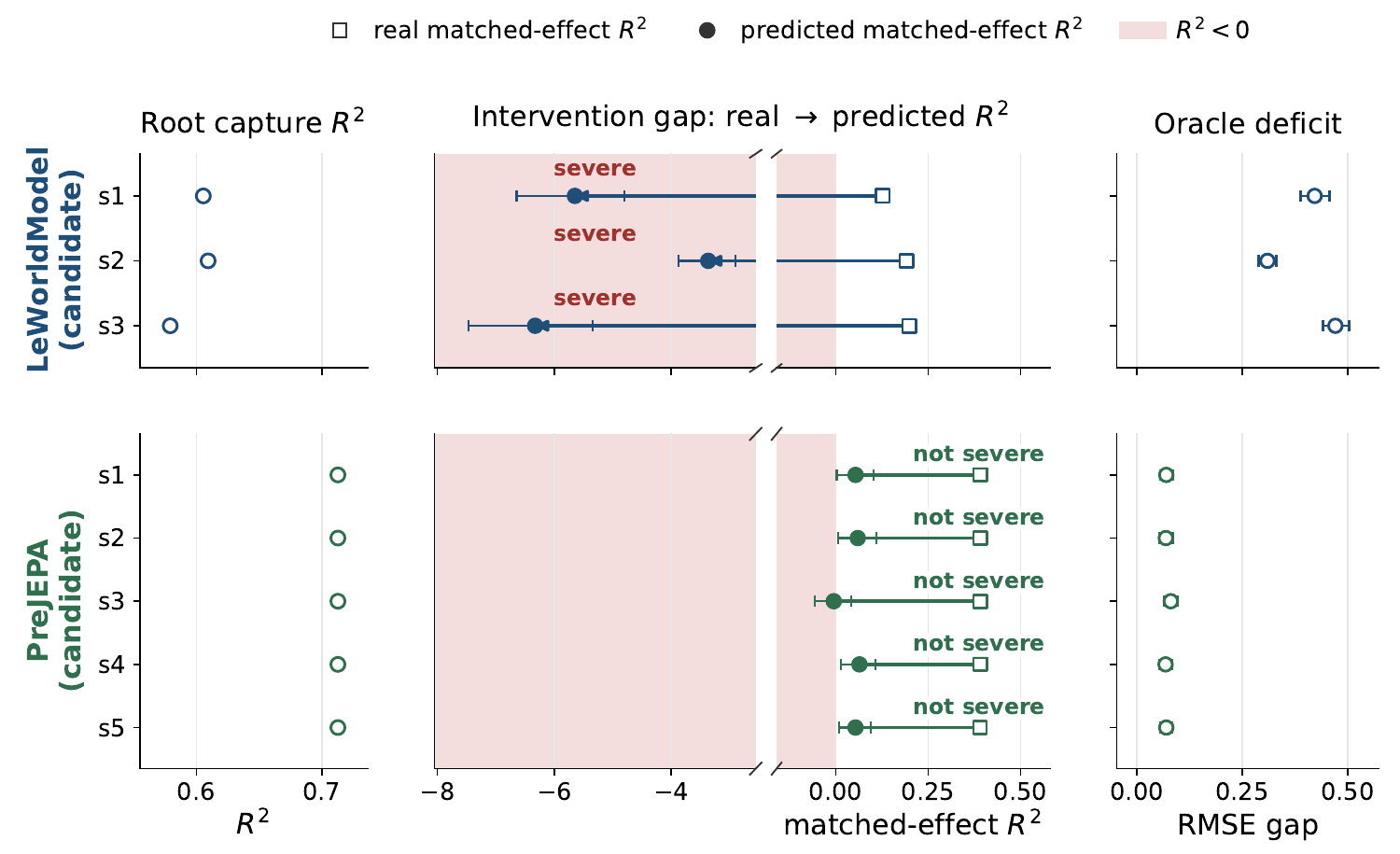}
\caption{\textbf{Intervention gap and non-replication on Cheetah.}
Rows are architectures. The left column is current-query capture \(R^2\).
The middle column is the matched-effect gap: open squares are real
matched-effect \(R^2\) and filled circles are predicted matched-effect \(R^2\)
(family 95\% intervals). Arrows run from the environment effect to the
model's predicted effect. That panel uses a broken \(R^2\) axis so both ends
remain readable; the omitted interval contains no plotted values, and the
horizontal scale is shared across architectures within each pane. The shaded
region is \(R^2<0\). Tags mark the frozen below-zero distortion gate:
severe means predicted \(R^2\) and its 95\% upper endpoint are both \(<0\);
not severe means that conjunction does not hold. The right column is the
oracle deficit (paired predicted-minus-oracle RMSE difference for
LeWorldModel; predicted-minus-oracle RMSE for PreJEPA); both \(>0\) means
worse than the environment-endpoint oracle. Candidate labels are not
comparable across rows. Intervals use 4{,}096 whole-family bootstrap
resamples (\(n=96\) evaluation families within each candidate). Readouts and
thresholds were frozen on calibration before evaluation. LeWorldModel shows
capture and real-effect resolution with predicted effects worse than zero and
worse than the oracle. PreJEPA retains capture, real-effect resolution, and
oracle deficit but does not reproduce the severe conjunction. Two
architectures do not license a population-of-architectures claim; PreJEPA
non-replication is not evidence of accurate propagation.}
\label{fig:intervention_gap_dissociation}
\end{figure}

Five independently initialized action-conditioned predictors over a shared
frozen DINOv2 backbone \citep{oquab2024dinov2} are evaluated with the same
four-part decision structure, with candidate selection completed before any
matched-intervention outcome is seen. Every predictor passes root capture,
real-effect resolution, and the oracle-relative comparison; none satisfies
the severe below-zero condition observed in LeWorldModel
(Table~\ref{tab:capture_propagation_summary};
Figure~\ref{fig:intervention_gap_dissociation}, bottom row). The severe
conjunction therefore does not replicate in a materially different visual
predictor, under a prespecified test capable of detecting it. Accurate propagation is not
established either: every predicted-minus-oracle error interval is strictly
positive, so all five imagined endpoints remain reliably worse than the real
endpoints they try to anticipate. Where LeWorldModel rotates and amplifies
the task effect, PreJEPA attenuates it toward zero: two different
geometries behind one shared deficit.

\subsection{A second task: heterogeneous severity on Finger Spin}
\label{sec:finger}

\begin{figure}[t]
\centering
\includegraphics[width=\linewidth]{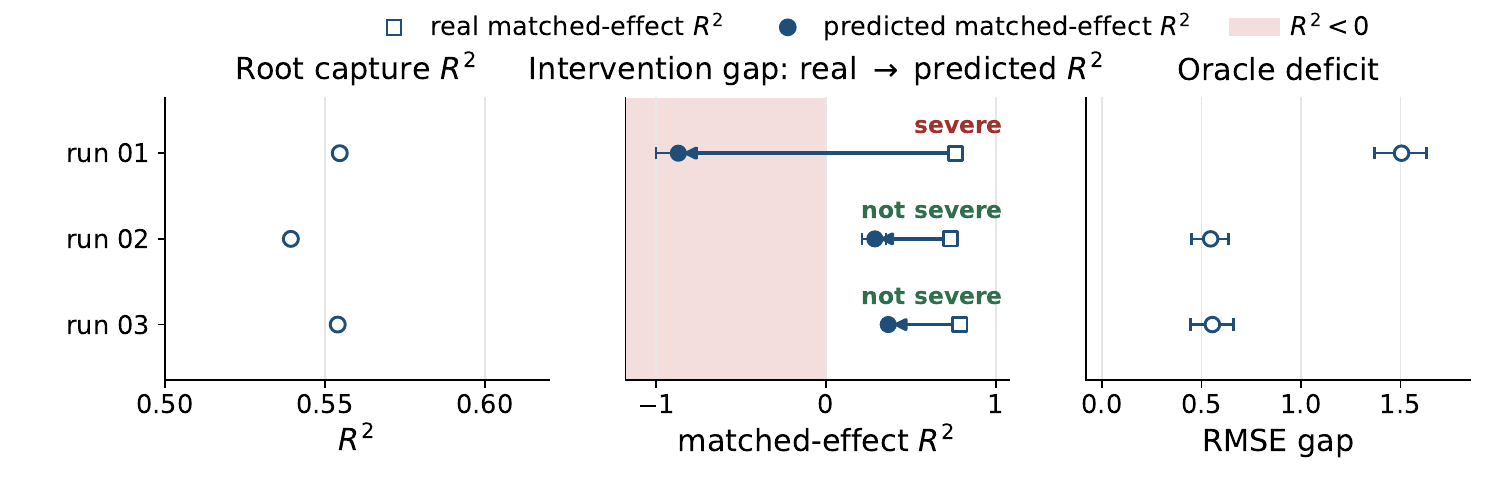}
\caption{\textbf{Finger Spin: oracle-relative gap with heterogeneous severity.}
Three independently seeded LeWorldModel runs on Finger Spin (endpoint
negative hinge velocity). Columns as in
Figure~\ref{fig:intervention_gap_dissociation}: capture \(R^2\);
matched-effect gap with the frozen severe/not-severe gate; paired
predicted-minus-oracle RMSE difference (\(>0\) means worse than the
environment-endpoint oracle). Intervals use 4{,}096 whole-family bootstrap
resamples (\(n=96\) evaluation families per candidate); readouts and
thresholds frozen on calibration before evaluation. Raw scores are not
comparable to Cheetah. The oracle-relative deficit replicates in every run;
severe distortion appears in one of three. One architecture and one
additional task; no universal mechanism, architecture ranking, or planner
claim.}
\label{fig:finger_dissociation}
\end{figure}

The same capture-gated protocol applied to Finger Spin, with endpoint negative
hinge velocity as the frozen query and three independently seeded LeWorldModel
runs, reproduces the structure of the dissociation beyond locomotion. All
three runs pass root capture and real-endpoint resolution, and every paired
predicted-minus-oracle interval is strictly positive. Severity, however, is
not seed-invariant: one run meets the severe below-zero certificate while
the other two retain positive but oracle-inferior predicted-effect signal
(Table~\ref{tab:capture_propagation_summary};
Figure~\ref{fig:finger_dissociation}). The oracle-relative gap replicates
across tasks; the extreme form does not replicate even across seeds of one
architecture on one task.

\subsection{Shared-bank geometry is candidate- and support-dependent}
\label{sec:geometry}

\begin{figure}[t]
\centering
\includegraphics[width=\linewidth]{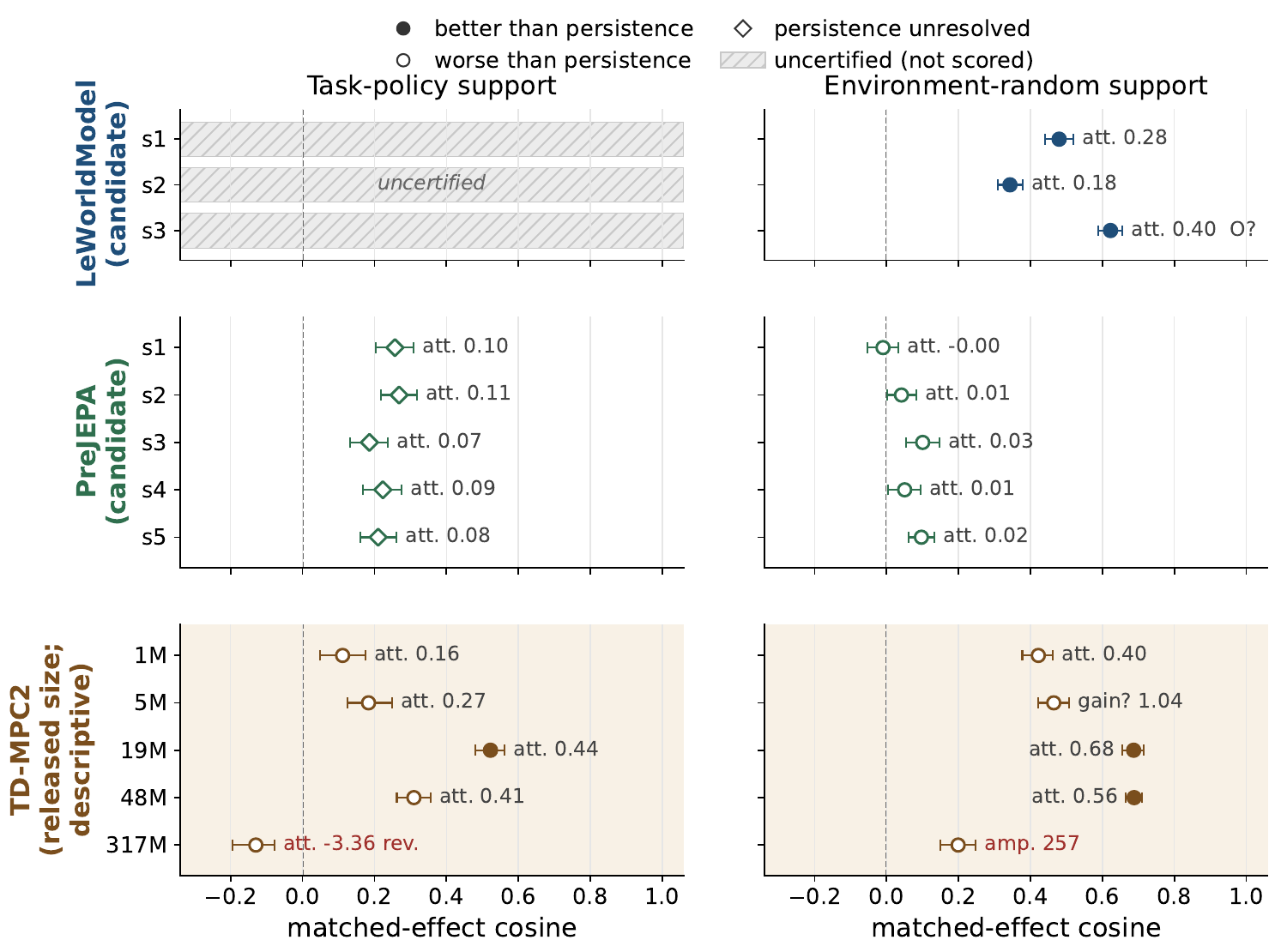}
\caption{\textbf{Shared-evaluation action-effect geometry.}
Rows are architectures and columns are intervention supports on identical
roots, five-step actions, and environment endpoints. Within each column
the horizontal axis is matched-effect cosine and is shared across
architectures; the vertical axis lists candidates and is not comparable across
rows (three LeWorldModel checkpoints, five PreJEPA seeds, five dependent
TD-MPC2 released sizes). Intervals use 4{,}096 whole-family bootstrap
resamples on the complete reset family (256 permutations). Hatched cells fail
real-effect resolvability or capture on this bank and are uncertified for
propagation; they are not scored as propagation successes or failures.
Numeric tags are frozen gain (att.~attenuation, amp.~amplification);
``rev.'' marks a negative-direction label; ``O?'' marks an unresolved oracle
comparison. Marker fill encodes the persistence comparison. Thirteen
primary-core cells meet inclusion criteria; every architecture-by-support
summary remains inconclusive. The TD-MPC2 row is descriptive only (dependent
released sizes). This figure does not rank architectures or establish a
size law.}
\label{fig:common_bank_geometry}
\end{figure}

The confirmatory rosters above use valid but different environment families.
A shared evaluation places the frozen LeWorldModel and PreJEPA candidates,
together with a descriptive panel of released TD-MPC2 models
\citep{hansen2024tdmpc2}, on identical roots, five-step actions, and
environment endpoints, with task-policy and environment-random supports
analyzed separately. No unanimous architecture signature emerges
(Figure~\ref{fig:common_bank_geometry}). LeWorldModel task-policy cells fail
real-effect resolution on this bank and remain uncertified for propagation;
PreJEPA task-policy effects are consistently attenuated; TD-MPC2 exhibits
large checkpoint-to-checkpoint and support variation, including a
sign-reversal and extreme-amplification pattern in its largest released
checkpoint across supports, with no monotonic size pattern. Every
prespecified architecture-by-support summary remains inconclusive.

Together, these boundary results support the following conclusion:
capture and propagation remain separable wherever tested, and an
oracle-relative propagation deficit has appeared in every qualified roster,
but the severity and geometry of the failure are properties of the
candidate, task, seed, and support, not of latent world models as a class.
The sharp Cheetah classification is valid on its frozen support and is not a
support-invariant property of the model.

\section{Auditing in Practice: Surfaces, Uncertainty, and Support}
\label{sec:practice}

If intervention fidelity is conditional and cannot be read off the value
channel, the practical question is how to audit it and when to trust the
model's own signals. Three results answer parts of that question: which
surface to read, how far disagreement can be trusted, and which gate binds
first.

\subsection{The query is carried by the posterior distribution, not its
sample}
\label{sec:surfaces}

Auditing presupposes a state surface that carries the query. In a
DreamerV3 study \citep{hafner2023dreamerv3}, six native surfaces were
localized for the current query across four training checkpoints under
frozen capture gates. Centered posterior logits are the sole surface passing
both gates at every checkpoint; the sampled categorical latent, its mode,
the deterministic recurrent state, and their concatenation pass at none, and
finite sample budgets through 1{,}024 samples do not bridge to the
exact-logit result (Figure~\ref{fig:dreamer_surface_grid};
Appendix~\ref{app:dreamer}). The information the audit
needs lives in the posterior \emph{distribution}, not in any single sampled
state the rollout machinery consumes. This is a representation-boundary
result: it licenses a distribution-level readout for auditing and does not
test Dreamer propagation, which remains uncertified.

\subsection{Disagreement ranks error only near support}
\label{sec:disagreement}

\begin{figure}[t]
\centering
\includegraphics[width=\linewidth]{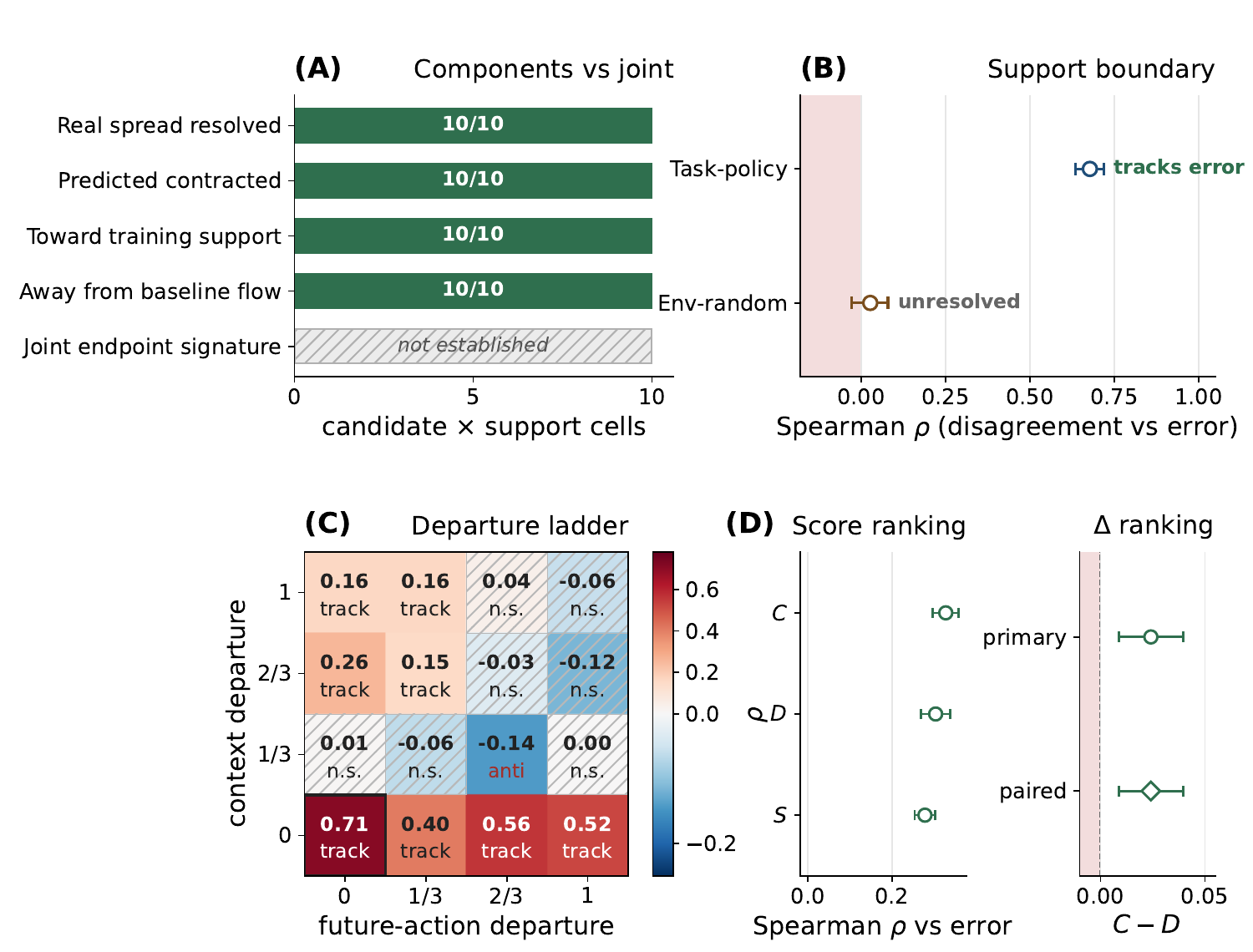}
\caption{\textbf{Support-linked signatures and departure ladder.}
(A--B)~PreJEPA endpoint/support analysis (five candidates; complete
environment family; 4{,}096 whole-family bootstraps). All ten
candidate--support cells resolve real branch spread, contract predicted
spread, move toward training support, and move away from same-support
baseline flow; the prespecified joint endpoint signature is not established.
Panel~B is family-level Spearman \(\rho\) between seed disagreement and
future-feature error under intervention (task-policy
\(\rho=0.678\), \([0.635,0.719]\); environment-random \(\rho=0.027\),
\([-0.029,0.080]\)).
(C--D)~Context--action support ladder (\(n=384\) evaluation families;
\(4\times4\) grid; 96 families per cell; primary render lane). Cell color is
Spearman \(\rho\); hatched cells are unresolved; the boxed corner is
near-policy support. Context and future-action components are each detectable
while their interaction is inconclusive.
\(C=0.5D_{\mathrm{pct}}+0.5S\) improves global ranking over disagreement
\(D\) (\(\Delta=0.0241\), 95\% interval \([0.0090,0.0397]\)); the paired
lane is a technical sensitivity, not an independent replication. This is
within-PreJEPA evidence only and does not calibrate epistemic uncertainty.}
\label{fig:support_ladder}
\end{figure}

Ensemble disagreement is the standard epistemic signal for learned dynamics
\citep{lakshminarayanan2017ensembles,chua2018pets}. Within the five-seed
PreJEPA family it behaves as a sharply support-relative diagnostic:
disagreement ranks future-feature error under task-policy support
(\(\rho=0.678\)) and is inconclusive under environment-random support
(\(\rho=0.027\)); a follow-up grid that independently varies root-context and
future-action departure shows the tracking deteriorating along both axes
(Figure~\ref{fig:support_ladder}). Predicted branches also contract relative
to encoded real branches and move toward the offline-training support while
moving away from same-support environment trajectories; the prespecified
joint attractor signature is not established. Within this family, a
prespecified equal-weight combination of disagreement percentile and support
score improves global error ranking over disagreement alone. None of this
calibrates epistemic uncertainty; it bounds where the proxy is informative.

\subsection{A frozen support correction fails held-out transfer in both
directions}
\label{sec:heldout}

The within-family success suggests a general rule: add a support term to
disagreement everywhere. A bidirectional held-out test falsifies that frozen
rule. The equal-weight score is frozen on PreJEPA and transferred without
refitting, first to three independently trained held-out TD-MPC2 runs, then
reciprocally, with TD-MPC2 joining development, to three independently
seeded held-out LeWorldModel runs on a fresh environment bank. Its
ingredients are the standard off-support penalties: native disagreement
\citep{yu2020mopo,kidambi2020morel} and a feature-space support distance
descended from Mahalanobis out-of-distribution scores
\citep{lee2018mahalanobis}.

\begin{figure}[t]
\centering
\includegraphics[width=\linewidth]{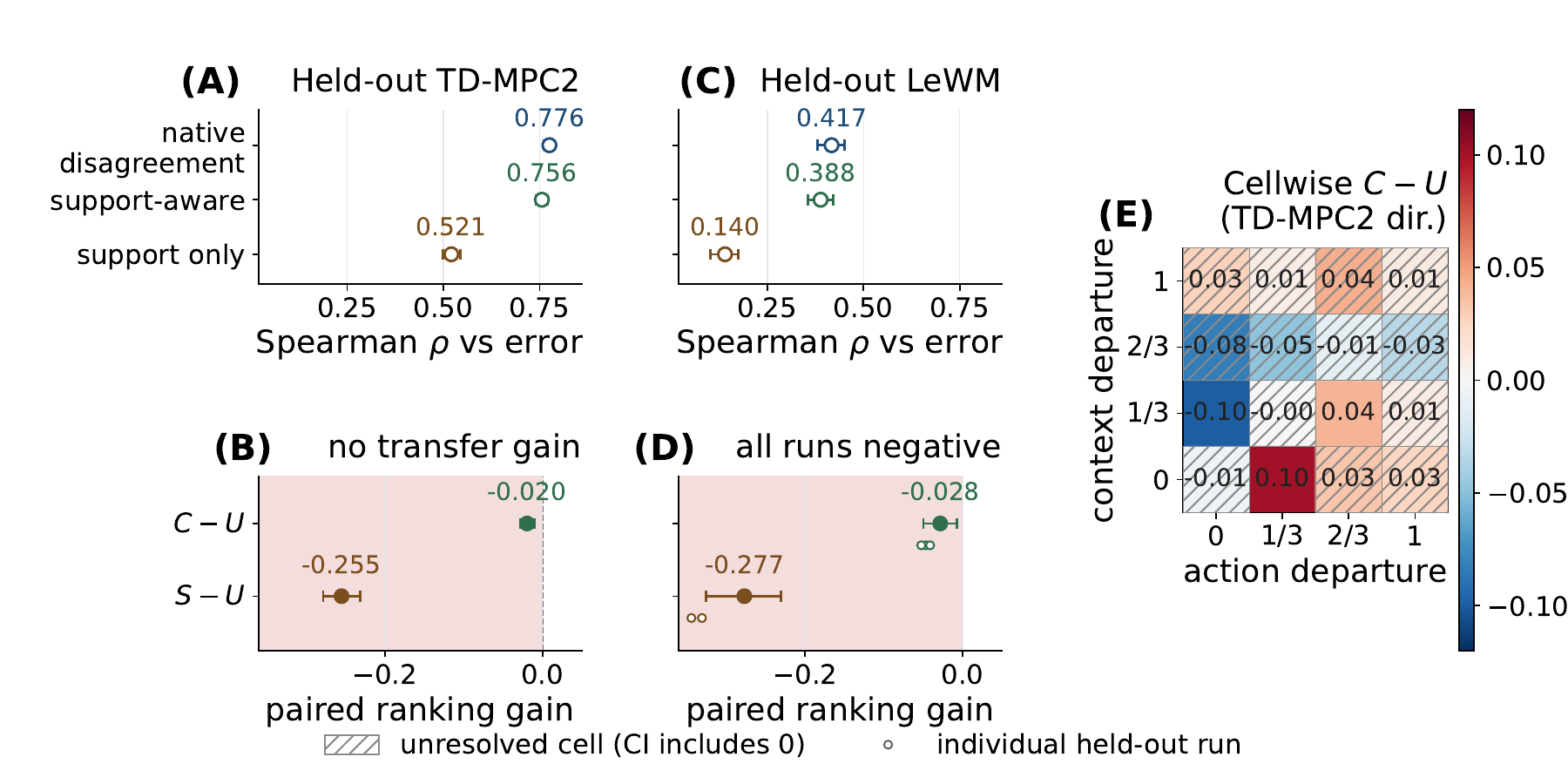}
\caption{\textbf{Frozen support augmentation fails held-out transfer in both
directions.}
The score \(C=0.5U_{\mathrm{pct}}+0.5S\) is frozen on PreJEPA and applied
without refitting.
(A--B)~Held-out TD-MPC2 (three independently trained 4M runs).
Complete-role Spearman \(\rho\) between each score and endpoint error:
native disagreement \(U\) \(0.776\) (\([0.762,0.789]\)); frozen combined
\(C\) \(0.756\) (\([0.741,0.770]\)); support distance alone \(0.521\)
(\([0.498,0.544]\)). Paired ranking gain relative to \(U\):
\(C-U=-0.020\) (\([-0.028,-0.010]\)); the shaded region is
\(\Delta\le 0\).
(C--D)~Reciprocal held-out LeWorldModel (three independently seeded
epoch-100 runs on a fresh environment bank). The pattern repeats at a lower
ranking level: \(U\) \(0.417\) (\([0.381,0.451]\)); \(C\) \(0.388\)
(\([0.355,0.420]\)); support alone \(0.140\); gain \(-0.028\)
(\([-0.050,-0.007]\)). Open circles are individual held-out runs; each has
a strictly negative combined-minus-native interval
(Appendix~\ref{app:candidate_results}).
(E)~Context \(\times\) action map of \(C-U\) in the TD-MPC2 direction.
Hatched cells have a 95\% interval that includes zero and are unresolved;
resolved cells remain mixed in sign and do not reverse the complete-role
label.
The \(\rho\) axes in A and C are shared so the lower reciprocal level is
visible; the two directions are analyzed separately and never pooled.
Intervals use 4{,}096 whole-family context-stratified bootstrap draws
(\(n=384\) evaluation families per direction; 4{,}608 context--action rows
as repeated measures). One task and bank design per direction; not
calibration, not a universal no-support law, and silent on reward, return,
and decision value.}
\label{fig:heldout_transfer}
\end{figure}

In both directions the same pattern holds: native disagreement ranks
endpoint error informatively, the frozen combined score is modestly but
reliably worse than disagreement alone, and support distance alone is much
weaker (Figure~\ref{fig:heldout_transfer};
Appendix~\ref{app:candidate_results}). In the reciprocal direction every
individual held-out run has a strictly negative combined-minus-native
interval. The two directions share one task and bank design, so they bound
this frozen additive rule rather than every support correction; they do not
establish a universal no-support law. The practice they support is
conservative: treat architecture-native disagreement as the default
transferable error-ranking signal, and treat any support augmentation as a
family-specific tool whose transfer must be revalidated before reuse.

\subsection{Capture is the first gate to fail in practice}
\label{sec:capturefirst}

The gate order of Section~\ref{sec:gatedaudit} is also where audits fail in
practice. In a matched-shell audit of three independently trained TD-MPC2
runs, every primary cell terminated at the first gates: absolute root
capture passed on task-policy roots but real-endpoint effect resolution
failed at both tested action radii, and on environment-random roots absolute
capture itself failed, so no propagation cell was interpretable in either
direction (Appendix~\ref{app:closure_boundary}). A readout qualified for
broad state reconstruction on one bank did not guarantee either support
transfer for the scalar query or sufficient resolution after
endpoint-minus-root differencing. The qualified environment-side closure
result shows the same limitation at the target level: closure held on matched
action shells, not on the full action ball.

The resulting protocol is brief. Declare the query, physical horizon, and
support before any candidate is evaluated. Qualify the readout twice, on
real roots and on real observed endpoints, per support. Interpret imagined
propagation only for cells that pass both, against persistence, zero effect,
and the environment-endpoint oracle. Read the model's native interface and,
where the state is distributional, the distribution rather than its sample.
Treat disagreement as informative only near training support, and require
any support correction to qualify on held-out families before use.
Fidelity, throughout, is a relation among model, query, horizon, support,
and interface, not a single score.

\section{Related Work}
\label{sec:related}

\paragraph{World models for planning.}
Learning a model of action consequences and planning inside it is a
long-standing program \citep{sutton1991dyna,ha2018worldmodels}. Modern latent
world models couple an encoder with an open-loop latent transition that a
planner or imagined-rollout learner consumes directly
\citep{hafner2019planet,hafner2020dreamer,hafner2021dreamerv2,hafner2023dreamerv3,schrittwieser2020muzero,hansen2022tdmpc,hansen2024tdmpc2}.
The audit evaluates exactly this planner-exposed interface: real encoded
roots and the model's own open-loop transitions.

\paragraph{Value equivalence and objective mismatch.}
Value-aware and value-equivalent model learning restrict model accuracy to
the policy and value classes relevant for planning
\citep{farahmand2017vaml,grimm2020value,grimm2021proper,silver2017predictron},
and one-step likelihood is known to correlate poorly with control performance
\citep{lambert2020objective}. Our distinction is operational: the evaluation
target is constructed outside candidate models, representational capture is
tested on real planning states, and transition propagation is interpreted
only after capture and persistence checks. The contribution is not the
existence of function-relative operator error, but a prospective measurement
protocol that separates target availability, real-effect resolution, and
fidelity of the model's own transitions. Section~\ref{sec:valuechannel} adds
a released-checkpoint instance of the mismatch: reward fit can be small and
nearly flat across checkpoints whose operator error and planning return vary
widely.

\paragraph{Rollout error and model exploitation.}
One-step-accurate learned models can compound error under the distribution
the planner induces \citep{ross2012agnostic,janner2019mbpo}, produce unstable
multi-step sample rollouts \citep{talvitie2014model,talvitie2017self}, and be
exploited by policy optimization in regions where they are wrong
\citep{kurutach2018model}. These accounts concern training and control
performance in aggregate; the present protocol instead measures the
propagation side directly, per query and per matched intervention, and only
after capture and real-effect prerequisites pass.

\paragraph{Evaluating world models beyond return.}
A growing line evaluates world models against structural criteria rather than
next-step loss or realized return. \citet{vafa2024evaluating} show that
generative sequence models can pass next-step diagnostics while their
implicit transition structure is incoherent, in a discrete-automaton setting;
the present audit is a continuous-control, intervention-matched analogue with
explicit capture gating and persistence and oracle comparators.
\citet{yu2026decision} argue, as a position, that world-model evaluation
should center interventional reasoning and rollout validity; we contribute an
executed, prospectively gated instance of such an audit together with its
boundary conditions. \citet{zhang2026probing} probe the latents of learned
environment simulators and intervene on latent directions; our interventions
are instead matched in the environment and scored at its executed endpoints.
Action-sensitivity metrics for generative interactive models
\citep{bruce2024genie,arai2024actbench} share the concern that learned models
under-use action inputs; the matched-contrast design tests this against real
action consequences rather than against alternative model rollouts.

\paragraph{Operator-theoretic language.}
Koopman and dynamic-mode methods lift dynamics to linear operators on finite
observable families and estimate them from data
\citep{koopman1931,schmid2010dmd,williams2015edmd,mezic2005,korda2018mpc,brunton2022koopman}.
We use that language to estimate a finite-horizon environment-side closure,
not to claim global linearity or spectral recovery.

\paragraph{Probing internal representations.}
Linear-probe methodology motivates explicit controls for readout capacity and
selectivity \citep{alain2017probes,hewitt-liang-2019-designing,belinkov2022probing}.
Probes have recovered linearly readable internal world state in sequence
models \citep{li2023othello,nanda2023emergent}, while amnesic probing shows
that readability does not imply behavioral use \citep{elazar2021amnesic}. The
capture gate is therefore paired with a separate propagation test rather than
read as evidence that decoded information is used.

\paragraph{Uncertainty and support.}
Ensemble disagreement is the standard epistemic signal for learned dynamics
\citep{lakshminarayanan2017ensembles,chua2018pets} and is used to penalize or
truncate off-support rollouts in offline model-based reinforcement learning
\citep{yu2020mopo,kidambi2020morel}. Its reliability degrades under
distribution shift \citep{ovadia2019trust}, and implemented disagreement
penalties track true model error inconsistently \citep{lu2022revisiting}.
Feature-space support distance descends from Mahalanobis
out-of-distribution scores \citep{lee2018mahalanobis}.
\citet{berger2026biased} report that ensemble disagreement in RSSM rollouts can
decrease while decoded physical discrepancy increases, and that prior
rollouts can overestimate reward relative to posterior-informed rollouts. This
external result motivates separating latent familiarity from trajectory
fidelity. It is not matched-intervention evidence for our candidates, and its
attractor account is not used as a causal explanation of our PreJEPA results.
Our support-relative findings are complementary: internal agreement or
proximity to familiar latent regions need not imply accurate physical action
effects, and a support-aware ranking score that helps within one family need
not transfer to another.

\paragraph{Self-supervised world models.}
Joint-embedding predictive architectures motivate reconstruction-free latent
world models \citep{lecun2022path,assran2023ijepa,balestriero2025lejepa}, and
latent dynamics over frozen self-supervised visual features support zero-shot
planning \citep{oquab2024dinov2,zhou2025dinowm}. The LeWorldModel and PreJEPA
candidates instantiate this family; the audit treats them as frozen
candidates rather than as representation or training proposals.

\section{Scope and Limitations}
\label{sec:limitations}

The learned-model evidence covers Cheetah locomotion and Finger Spin, each
with one primary scalar query and a five-step physical horizon. An
eligibility survey examined 32 systems spanning 13 plausible families; only
the five-run PreJEPA family passed every original primary criterion. Released
TD-MPC2 sizes and some LeWorldModel checkpoints remain dependent descriptive
panels. The evidence therefore supports neither cross-family inference nor a
population-level architecture ranking. A Walker continuation is authorized
but not yet terminal, and cannot be pooled with Cheetah or Finger.

The Section~\ref{sec:valuechannel} results are correlational and
single-environment. The size-sweep association uses five dependent released
checkpoints from one training recipe; it is a flag, not a selection rule.
The anchoring comparison spans two model families with one anchored seed,
differing input modalities, and differing native prediction modes; it bounds
the claim that task anchoring secures operator fidelity and supports no
architecture ranking. Raw scores from the coarse operator diagnostic and the
capture-gated audit are not comparable.

Capture is relative to the fixed readout family and does not establish
absence or availability under every nonlinear decoder. Probe success also
does not show that a planner uses the decoded information
\citep{elazar2021amnesic}. Propagation fidelity need not imply high return
when planning is discontinuous or action margins are small; the experiments
provide no checkpoint-selection result, planning-return improvement, or
return guarantee.

The real-model results use literal task queries. The environment-side
closure is calibrated synthetically, qualifies only on matched action shells,
and its candidate-model decision value remains untested: the first model-side
shell audit terminated at the capture gates
(Appendix~\ref{app:closure_boundary}).

The support-reference analyses remain correlational. Endpoint contraction,
motion toward training support, and disagreement--error association do not
identify a causal mechanism, and the geometric characterizations of
Sections~\ref{sec:geometryoffailure} and~\ref{sec:boundaries} (rotation
with excess gain in one family, attenuation in another) constrain candidate
mechanisms without selecting one. External evidence that latent rollouts can
drift toward familiar regions while physical error grows
\citep{berger2026biased} motivates a support-attraction account but was not
obtained under matched interventions and is not used as an explanation here.
The held-out transfer results bound one frozen additive support rule on one
task and bank design; they do not bound every support correction, and the
Dreamer studies establish representation boundaries rather than a
propagation result, so Dreamer propagation remains uncertified.

\section{Conclusion}
\label{sec:conclusion}

Planning-time intervention fidelity is a distinct, measurable property of a
learned world model. In the settings tested here it is neither revealed by
reward fit nor ensured by task-anchored training; it can fail severely while
current-state capture, real-effect resolvability, and overall feature
alignment all pass; and its severity and geometry are properties of the
candidate, task, seed, and support rather than of latent world models as a
class. The audit that exposes this treats fidelity as a relation among the
model, task query, horizon, intervention support, and native interface:
qualified capture-first, read on the surface that actually carries the
query, with the model's own uncertainty trusted only near its training
support. Two questions are deliberately left open: which mechanism
produces rotation with excess gain in one family and attenuation in another,
and whether the audit's rankings carry decision value for checkpoint
selection and return. Both require experiments this framework now makes
well-posed.

\clearpage
\appendix
\section{Finite-Horizon Decomposition}
\label{app:decomposition}

The decomposition in Section~\ref{sec:gatedaudit} follows by inserting the
environment-side closure projection after each model transition, separating
the resulting one-step model and closure-approximation residuals, and
recursively applying the local stability bound. Its terms distinguish
root-capture error, transition error accumulated along the rollout, and target
approximation error. No term controls policy discontinuity or a planner's
action margin, so the bound is not a guarantee about realized return.

\section{Synthetic Calibration Suite}
\label{app:synthetic}

The synthetic suite contains ten cases: exact capture and propagation; exact
capture with incorrect dynamics; absent target information; slow persistence;
rank deficiency; shuffled actions; stochastic conditional expectations;
off-manifold deterministic means; harmless orthogonal basis changes; and
forbidden split overlap. The complete suite passes. These cases validate
the procedure and stopping rules, not the adequacy of a learned model or the
incremental value of empirical closure.

\section{Per-Candidate Results}
\label{app:candidate_results}

\begin{table}[h]
\centering
\caption{LeWorldModel confirmatory evaluation. Intervals use complete
environment-family bootstrap resamples. The final two columns report the upper
endpoint for predicted-effect \(R^2\) and the lower endpoint for
predicted-minus-oracle family RMSE.}
\label{tab:lewm_primary}
\begin{tabular}{lrrrrr}
\toprule
Checkpoint & Root \(R^2\) & Real effect \(R^2\) & Predicted effect \(R^2\)
& \(R^2\) upper & Error diff.\ lower \\
\midrule
s1 & 0.605 & 0.127 & -5.646 & -4.795 & 0.388 \\
s2 & 0.609 & 0.191 & -3.364 & -2.902 & 0.289 \\
s3 & 0.579 & 0.198 & -6.326 & -5.349 & 0.440 \\
\bottomrule
\end{tabular}
\end{table}

Each LeWorldModel checkpoint uses 96 independent evaluation environment
families, with three roots and six matched actuator contrasts per family.
Checkpoints remain separate descriptive members of one architecture.

\begin{table}[h]
\centering
\caption{PreJEPA prospective architecture contrast. Candidate rows remain
separate in the source results; the table records only architecture-wide
outcomes that hold for all five seeds.}
\label{tab:prejepa_primary}
\begin{tabular}{lcccc}
\toprule
Candidates & Root capture & Real effect \(R^2\) & Predicted effect \(R^2\)
& Oracle deficit \\
\midrule
5 seeds & 5/5 pass & \(\approx 0.390\) & \(-0.005\) to \(0.064\)
& 5/5 intervals \(>0\) \\
\bottomrule
\end{tabular}
\end{table}

None of the five PreJEPA candidates passes the strict below-zero distortion
condition. The severe LeWorldModel joint failure therefore does not replicate,
while accurate propagation is also not established.

\begin{table}[h]
\centering
\caption{Finger Spin LeWorldModel evaluation on the complete surface.
Intervals use 4{,}096 whole-family bootstrap resamples on 96 evaluation
families. The query is endpoint negative hinge velocity. Raw scores are not
comparable to Cheetah torso speed.}
\label{tab:finger_primary}
\setlength{\tabcolsep}{4pt}
\begin{tabular}{lrrrrr}
\toprule
Run & Root \(R^2\) & Real \(R^2\) & Pred.\ \(R^2\)
& Pred.\ 95\% & Oracle-err.\ 95\% \\
\midrule
01 & 0.555 & 0.763 & -0.869 & \([-1.003,-0.760]\) & \([1.368,1.632]\) \\
02 & 0.539 & 0.733 & 0.287 & \([0.212,0.352]\) & \([0.451,0.637]\) \\
03 & 0.554 & 0.789 & 0.367 & \([0.338,0.391]\) & \([0.447,0.659]\) \\
\bottomrule
\end{tabular}
\end{table}

Run 01 meets the severe below-zero certificate. Runs 02 and 03 retain
non-severe positive predicted-effect signal. All three have a strictly
positive oracle-relative deficit.

\begin{table}[h]
\centering
\caption{Bidirectional held-out support-augmentation ranking. Spearman
\(\rho\) uses 384 evaluation families per direction and 4{,}096
context-stratified whole-family bootstrap draws; the reciprocal direction
uses fresh environment-family identities. \(C\) is the frozen equal-weight
combination of native disagreement percentile and support distance; \(U\) is
native disagreement; \(S\) is support distance alone. Directions are never
pooled.}
\label{tab:heldout_primary}
\begin{tabular}{lrrrr}
\toprule
& \multicolumn{2}{c}{Held-out TD-MPC2 (4M)}
& \multicolumn{2}{c}{Held-out LeWM (epoch 100)} \\
\cmidrule(lr){2-3}\cmidrule(lr){4-5}
Score & \(\rho\) & 95\% interval & \(\rho\) & 95\% interval \\
\midrule
Native disagreement \(U\) & 0.776 & \([0.762,0.789]\)
  & 0.417 & \([0.381,0.451]\) \\
Support-aware \(C\) & 0.756 & \([0.741,0.770]\)
  & 0.388 & \([0.355,0.420]\) \\
Support only \(S\) & 0.521 & \([0.498,0.544]\)
  & 0.140 & \([0.102,0.175]\) \\
\(C-U\) & -0.020 & \([-0.028,-0.010]\)
  & -0.028 & \([-0.050,-0.007]\) \\
\(S-U\) & -0.255 & \([-0.278,-0.231]\)
  & -0.277 & \([-0.326,-0.231]\) \\
\bottomrule
\end{tabular}
\end{table}

\begin{table}[h]
\centering
\caption{Per-run reciprocal-direction ranking on held-out LeWorldModel.
Each run is one independently seeded epoch-100 candidate; intervals use the
same context-stratified whole-family bootstrap.}
\label{tab:heldout_reciprocal_runs}
\begin{tabular}{lrrr}
\toprule
Run & \(\rho_U\) & \(\rho_C\) & \(C-U\) (95\% interval) \\
\midrule
01 & 0.483 & 0.436 & \(-0.047\) \([-0.067,-0.026]\) \\
02 & 0.505 & 0.453 & \(-0.052\) \([-0.072,-0.033]\) \\
03 & 0.472 & 0.431 & \(-0.041\) \([-0.062,-0.021]\) \\
\bottomrule
\end{tabular}
\end{table}

Every held-out LeWorldModel run independently has a strictly negative
combined-minus-native interval, so the reciprocal complete-role label is not
driven by a single seed.

On the shared evaluation, 13 primary combinations meet the inclusion
criteria. Candidate and support rows remain separate because every
architecture-by-support summary is inconclusive. The support-departure
evaluation uses 384 independent families over a \(4\times4\) context--action
grid; its two technical replicate runs are paired repeats.

\section{Probe and Normalization Controls}
\label{app:controls}

Target construction, probe training, calibration, and evaluation use disjoint
environment-family identities. Target normalization and rank decisions use
development data only. Candidate models are unavailable during target
construction. Primary real-model tests use family-preserving permutations and
4,096 whole-family bootstrap resamples. Controls include label and feature
shuffles, horizon-zero identity, persistence or zero effect,
environment-endpoint decoding, wrong action, action null, cyclic actuator,
sign swap, and direct feature alignment where defined.

\section{Dreamer Interface Boundaries}
\label{app:dreamer}

\begin{figure}[t]
\centering
\includegraphics[width=0.85\linewidth]{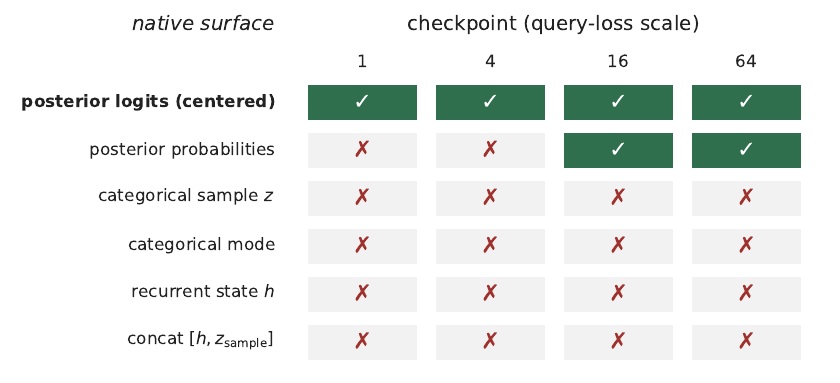}
\caption{\textbf{Dreamer current-query surface localization.}
Frozen capture-gate outcomes for six native surfaces across the four
held-out checkpoints (columns indexed by query-loss scale). Centered
posterior logits pass both coordinate gates at every checkpoint; posterior
probabilities pass only at scales 16 and 64; the sampled categorical state,
its mode, the deterministic recurrent state, and the concatenation pass
none. Pass/fail is encoded redundantly by color and glyph. This is
current-state availability only; propagation is not tested.}
\label{fig:dreamer_surface_grid}
\end{figure}

The current-state localization study evaluates four checkpoints, permanently
held out of later tests, on 64 fresh validation environment families and 2,048
starting states per checkpoint and representation. Centered posterior logits
are the sole
surface passing both current-query coordinate gates at every checkpoint.
Single sampled categorical states, categorical modes, deterministic recurrent
states, and the tested recurrent-state/sample concatenation pass none;
posterior probabilities pass only at query-loss scales 16 and 64
(Figure~\ref{fig:dreamer_surface_grid}). This shows
that current information is available after the fact, not that it is
propagated.

A separate paired study transfers the centered-logit instrument to two
independent \texttt{free\_nats} pairs and evaluates categorical budgets from
one through \(1024\) samples. Exact-logit instruments pass in all four runs,
but no sampled budget passes both coordinates for all runs despite monotone
approximation improvement. The intended tight-minus-reference raw-KL
direction holds in all training intervals, while both terminal full-prefix
contrasts reverse, to \(+1.406885\) and \(+0.609669\); the corresponding
one-nat occupancy comparisons are zero. The finite-sample argument and
terminal raw-KL direction are
therefore not established, and propagation is not tested.

\section{Empirical-Closure Qualification Boundary}
\label{app:closure_boundary}

The original Cheetah closure qualification uses 48 development and 48
validation families per support, with supports kept separate. Both supports
return no leading component above the significance threshold, and rank zero, so
all later
target constructions, probes, candidate scores, and
evaluation assignments remain unused.

A prospective geometry diagnosis establishes why this rank-zero result cannot
be interpreted as absence of action-conditioned structure. The action slots
were sampled independently within each family and were therefore exchangeable
identifiers rather than common semantic interventions. Action-effect magnitude
and the ordering from \(H=1\) through \(H=5\) persist, but common linear
predictability is weak and physical dimensionality is not identifiable from
the slot SVD. The original rank estimate cannot be used. A valid
redesign requires a fresh shared semantic actuator--time dictionary rather
than a relaxed rank threshold.

\begin{figure}[t]
\centering
\includegraphics[width=\linewidth]{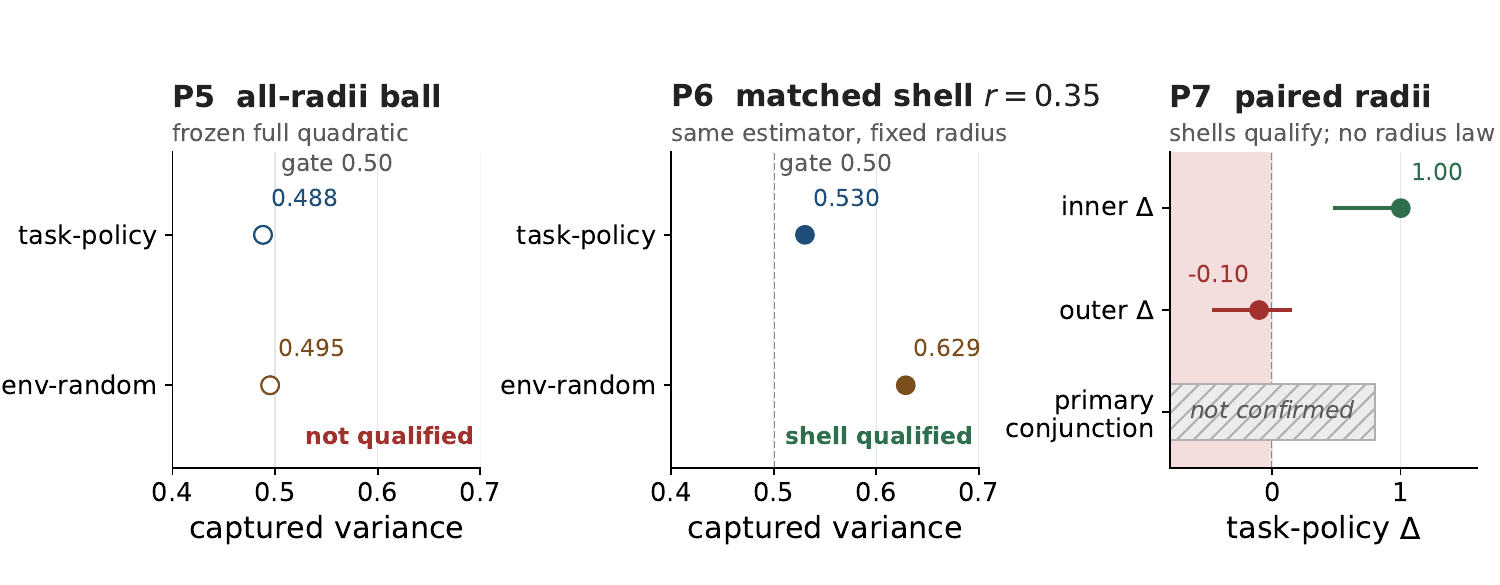}
\caption{\textbf{Environment-side shell localization.}
Successive C2 terminals on Cheetah (E60--E63). The frozen full-quadratic
estimator fails the all-radii action-ball gate on both supports (P5) and
qualifies on the matched \(r=0.35\) shell on both supports (P6). A paired
inner/outer shell test again qualifies both shells (P7), but the primary
task-policy radius-conditioning contrast is not confirmed. Supports are
analyzed separately; environment-random cannot rescue the primary arm. This
is environment-side estimand localization only: it does not establish
candidate-model capture, propagation, decision validity, regret, or return.}
\label{fig:c2_shell_localization}
\end{figure}

Subsequent environment-side terminals localize that redesigned estimand
(E60--E63). The frozen full-quadratic estimator fails over the all-radii
action ball and qualifies on a matched fixed-radius shell; a paired-radius
test again qualifies both shells but does not confirm task-policy radius
conditioning
(Figure~\ref{fig:c2_shell_localization}).

The first model-side matched-shell audit applies the qualified shells to
three independently trained TD-MPC2 runs with paired 1M/4M snapshots, two
supports, and two radii (24 cells). Every primary cell terminates at the
capture gates: on task-policy roots, absolute root capture passes for all
three 4M runs while paired real-endpoint action-effect capture fails at both
radii; on environment-random roots, absolute capture itself fails for all
three runs. Each 1M cell receives the same label as its paired 4M cell, and
all truth-fidelity, oracle-relative, and imagined-geometry labels are
uncertified by capture failure. A readout qualified for broad 17-dimensional
state reconstruction on a disjoint development bank guaranteed neither
support transfer for the scalar query nor sufficient resolution after
endpoint-minus-root differencing; absolute-query capture and paired-effect
resolution are distinct instrument requirements.

\section{Inferential Breadth}
\label{app:breadth}

The eligibility survey examines 32 systems spanning 13 plausible
architecture families. Only the five-run PreJEPA family satisfies every
original primary criterion; four runs are primary and one is an extra run.
Five released
TD-MPC2 sizes remain descriptive examples rather than independent training
runs. The survey therefore
supports no cross-family inference, monotonic size law, or model ranking.

The separate exploratory TD-MPC2 panel contains 80 complete scores.
The 317M model combines accurate capture with extreme predicted
action-effect error. A 19M model on policy support outperforms persistence,
while the same candidate is worse under environment-random starting states.
These contrasts establish
support sensitivity within a dependent released-size panel, not confirmatory
architecture evidence.

\end{document}